# MOVIE RECOMENDATION SYSTEMS USING AN ARTIFICIAL IMMUNE SYSTEM

Q Chen, U Aickelin,
qxc@cs.nott.ac.uk, uwe.aickelin@nottingham.ac.uk,
School of Computer Science and IT,
University of Nottingham, UK

## ABSTRACT

We apply the Artificial Immune System (AIS) technology to the Collaborative Filtering (CF) technology when we build the movie recommendation system. Two different affinity measure algorithms of AIS, Kendall tau and Weighted Kappa, are used to calculate the correlation coefficients for this movie recommendation system. From the testing we think that Weighted Kappa is more suitable than Kendall tau for movie problems.

## 1. INTRODUCTION

In everyday life, we often face a situation in which we need to make choices without sufficient personal experience. These arising needs call for effective recommendation systems to assist us in making these choices. Today's techniques used in recommendation systems are mainly collaborative filtering technology and content-based technology [1]. Collaborative filtering technology implied with AIS is used in our project. Two correlation methods, Weighted Kappa and Kendall tau, are used to calculate the correlation coefficients, and their results are compared.

### Collaborative Filtering Technology (CF)

CF is the technology offering users recommendations by getting recommendations from the people who have similar preferences with the users [2]. The collaborative filtering technology can offer you recommendations to items even though you do not know the content of these items. That is a big advantage. The group of people with similar preferences with the user entirely determine the predictions of the user who request recommendation, so it is very important to choose the group of people.

### Human Immune System (HIS)

HIS is the defence system of our body which can protect our body against infections [4]. The antigens (Ag) attacking our body can stimulate the immune system to produce antibodies.

### Artificial Immune Systems (AIS)

AIS are distributed and adaptive systems using the models and principles derived from the Human Immune System, AIS are used for problem solving. [3]

## 2. IMPLEMENTATION

Two movie recommendation systems are implemented using the collaborative filtering technology and AIS, one uses the Weighted Kappa method to calculate correlation coefficients, and the other uses the Kendall tau method. The AIS is built to select the group of people with similar movie preferences as the target. The people in the database are viewed as candidate antibodies, and the user who uses the movie recommendation system is viewed as an antigen. The two correlation methods are used to calculate the correlations between the antigen and antibodies, and the correlations between antibodies and antibodies.

### System Process

The figure below describes how our recommendation system works:

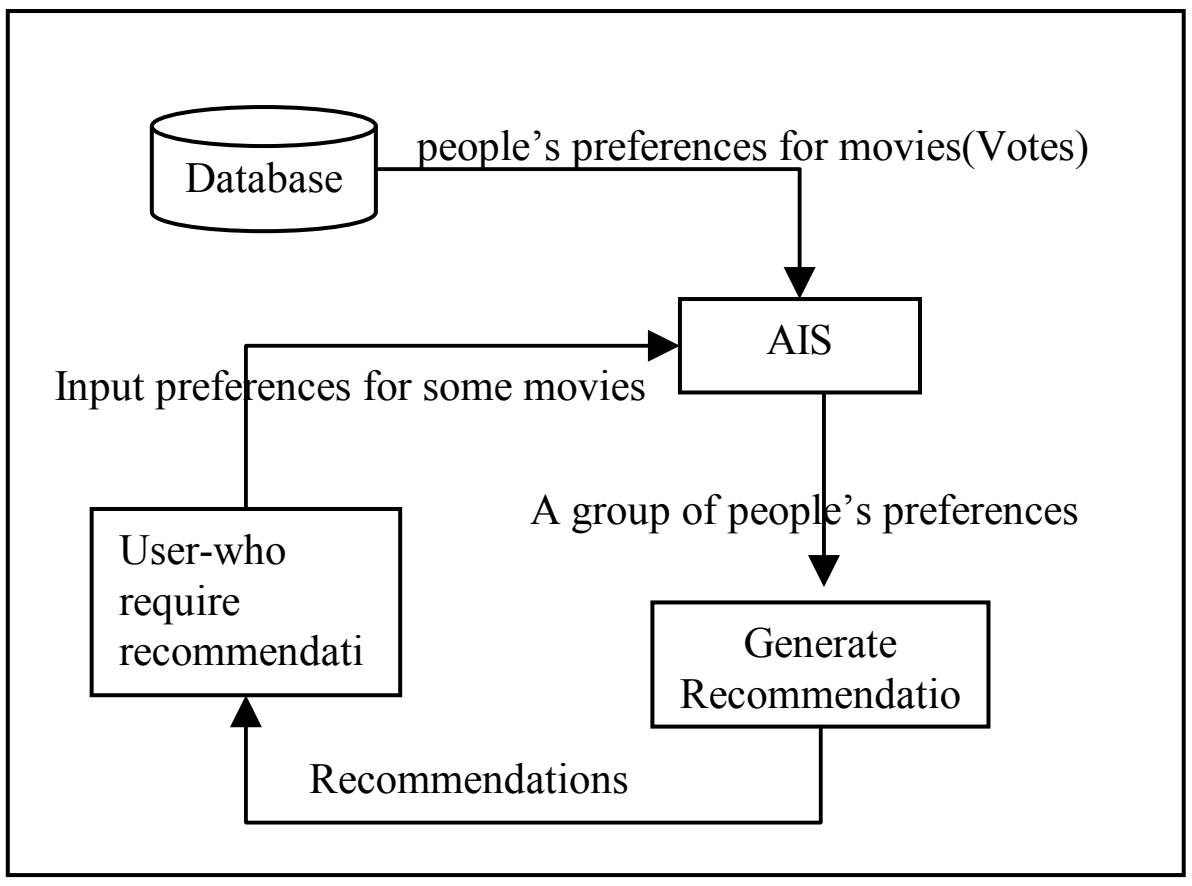

**Figure 1** (System process diagram)

1: There are some people's preferences stored in the database;
2: User inputs his preferences for the movies, and requires recommendations on some movies that he has not seen
3: AIS selects a group of people who have similar preferences with the user

4: The weighted average of the preferences for that group of people is calculated by the CF to generate recommendations which the user requires

**Immune Network Model**

The AIS model used in this project is the Immune Network Model built by Farmer et al [5] and modified by Cayzer et al [6]. This model is controlled by the Equation (1) which describes how the antibody's concentration changes. It increases for the antibody's matching to the antigen, and decreases for the antibody's matching to the other antibodies. And there exist the death rate, so if the antibody is neither bad nor good, its concentration also will decrease. When an antibody's concentration is below a value, we will delete it from the AIS, and choose another new antibody randomly from the database. When all antibodies in the AIS satisfy our requirement or there are no more antibodies can be chosen from the database, updating the AIS process will stop.

$$\frac{dx_i}{dt} = k_1 m_i x_i y - \frac{k_2}{n}\sum_{j=1}^{N} m_{i,j} x_i x_j - k_3 x_i$$

(**Equation 1** )

$y$ represents the concentration of antigen

$x_i$ represents the concentration of antibody i, $x_j$ represent the concentration of antibody j.

$m_{i,j}$ represents the affinity between the antibody i and j,

$m_i$ represents the affinity between the antibody i and the antigen

**Affinity Measure Algorithms**

We will use two different algorithms to calculate the affinity (correlation coefficient) - Kappa and Kendall tau. We will use this example below to explain how Weighted Kappa and Kendall tau work.

Example 1: (two person from the database, one's person Id is 50, the other's person Id is 70, {(movie-id$_1$, movie-vote$_1$); ……; (movie-id$_n$, movie-vote$_n$ )} is used to encode a person who has voted n movies. the vote scores is 0, 0.2, 0.4, 0.6, 0.8, 1.)

Person1 (id=50): { ***(2,1);*** (4,1); ***(19,0.6);*** (21,0.2); ***(24,0.8);*** (27,1); (31,1); ***(32,0.8);*** ***(62,1);*** ***(65,0.8);*** (76,1); (93,0.6); (94,0.8)}

Person2(id=70): {(1,0.8); ***(2,0.6);*** (5,0.6); (8,0.4); (13,0.2); (15,0); ***(19,0.2);*** ***(24,0.6);*** (25,0.4); ***(32,0.8);*** (34,0.8); (52,0.6); ***(62,0.8);*** ***(65,0);*** (70,0.6); (86,0.4); (87,0.2); (95,0.8); (107,0.6)}

The votes in bold indicate they are the votes for the movies which the two persons have seen in common.

**Weighted Kappa Algorithm** [8]: Weighted kappa is a method of calculating affinity (correlation coefficient), which is calculated using the Equation (2) from the observed and expected frequencies.

$$k_w = \frac{p_{o(w)} - p_{e(w)}}{1 - p_{e(w)}} \quad \textbf{Equation 2}$$

$P_{o(w)}$ represents the observed agreements

$P_{e(w)}$ represents the expected agreements by chance.

In the movie recommendation systems, all the persons in the person database chose the movies they had seen from the movie database, and ranked them. No agreements by chance exist, so $P_{e(w)}$ =0 and $k_w = P_{o(w)}$ . The $P_{e(w)}$ is calculated by the Equation (3).

$$p_{o(w)} = \frac{1}{n}\sum_{i=1}^{g}\sum_{j=1}^{g} w_{ij} f_{ij} \quad \textbf{Equation 3}$$

*g represents category.*

*n* represents the number of the observations in *g* categories.

$f_{ij}$ represent the number of agreements for the cell in row *i* and column *j*.

*wij* represents the weight value for the cell in row *i* and column *j* .

In our project, the category *g*=6 (for the user has 6 optional movie vote to choose from—0, 0.2, 0.4, 0.6, 0.8, 1); the observations number *n* is the number of movies two persons have seen in common, $w_{ij}$ can be got by the Equation (4).

$$w_{ij} = 1 - \frac{|i-j|}{g-1} \quad \textbf{Equation 4}$$

The difference between *i* and *j* is bigger, the weight is smaller; and the difference between *i* and *j* is smaller, the weight is bigger. When *i*=*j*, the weight will reach the biggest value 1.

Using the equation (4) we get the Table 1 below, which shows the weight values $w_{ij}$ for our project.

| | j =1 | j =2 | j =3 | j =4 | j =5 | j =6 |
|---|---|---|---|---|---|---|
| i = 1 | 1 | 0.8 | 0.6 | 0.4 | 0.2 | 0 |
| i = 2 | 0.8 | 1 | 0.8 | 0.6 | 0.4 | 0.2 |
| i = 3 | 0.6 | 0.8 | 1 | 0.8 | 0.6 | 0.4 |
| i = 4 | 0.4 | 0.6 | 0.8 | 1 | 0.8 | 0.6 |
| i = 5 | 0.2 | 0.4 | 0.6 | 0.8 | 1 | 0.8 |
| i= 6 | 0 | 0.2 | 0.4 | 0.6 | 0.8 | 1 |

**Table 1** (weight values table)

In example 1, the person 1 and person 2 have seen 6 movies in common, so the observations number is 6; they are movie 2, movie 19, movie 24, movie 32, movie 62, and movie 65.

For movie 2 person 1 vote it as "1"( row 6), person 2 Vote as "0.6"( column 4);

for movie 19 person 1 vote it as "0.6"(row 4), person 2 Vote as "0.2"( column 2);

for movie 24 person 1 vote it as "0.8"(row 5), person 2 Vote as "0.6" (column 4);

for movie 32 person 1 vote it as “0.8” (row 5), person 2 Vote as “0.8” (column 5);
for movie 62 person 1 vote it as “1” (row 6), person 2 Vote as “0.8” (column5);
for movie 65 person 1 vote it as “0.8” (row 5), person 2 Vote as “0”(column 1);
so we got the $f_{ij}$ (the number of agreements for the cell in row *i* and column *j*)

| 2 / 1 | 1 (0) | 2 (0.2) | 3 (0.4) | 4 (0.6) | 5 (0.8) | 6 (1) | Total |
|---|---|---|---|---|---|---|---|
| 1(0) | 0 | 0 | 0 | 0 | 0 | 0 | 0 |
| 2(0.2) | 0 | 0 | 0 | 0 | 0 | 0 | 0 |
| 3(0.4) | 0 | 0 | 0 | 0 | 0 | 0 | 0 |
| 4(0.6) | 0 | 1 | 0 | 0 | 0 | 0 | 1 |
| 5(0.8) | 1 | 0 | 0 | 1 | 1 | 0 | 3 |
| 6 (1) | 0 | 0 | 0 | 1 | 1 | 0 | 2 |
| Total | 1 | 1 | 0 | 2 | 2 | 0 | **6** |

**Table 2** (Agreement ($f_{ij}$) Table for example 1)

Weighted kappa correlation between person 2 and 1 is:
$k_{(w)}$=1/6*(0.2*1+0.6*1+0.8*1+0.6*1+0.8*1+1) =0.667
We know that the agreement between these two persons is good for the Table3 below.

| Value of kappa | Value of Kendall tau | Strength of agreement |
|---|---|---|
| <0.20 | -1~ -0.2 | Poor |
| 0.21-0.40 | -0.6 ~ -0.2 | Fair |
| 0.41-0.60 | -0.2 ~ 0.2 | Moderate |
| 0.61-0.80 | 0.2 ~ 0.6 | Good |
| 0.81-1.0 | 0.6 ~ 1 | Very good |

Table 3 (The table of kappa meaning) [6]

**Kendall tau** [9]: Kendall tau is another method of calculating affinity (correlation coefficient).
For a pair of observations ($X_i$, $Y_i$) and ($X_j$, $Y_j$), we view it as a concordant pair if $X_j$-$X_i$ and $Y_j$-$Y_i$ have the same sign; we view it as a discordant pair if $X_j$-$X_i$ and $Y_j$-$Y_i$ have opposite signs. *C* represents the number of concordant pairs; *D* represents the number of discordant pairs. Then we get the Equation (5), which is used to calculate the affinity (correlation coefficient).

$$\hat{\tau} = \frac{2S}{n(n-1)} , -1 \le \hat{\tau} \le +1 . \quad \textbf{(Equation 5)}$$

$\hat{\tau}$ represents Kendall tau affinity (correlation coefficient)
n represents observations number
*S* represents Kendall S, defined as: $S = C-D$
For *n* observations, there are ½*n*(*n*-1) pairs, either concordant or discordant. If they are all concordant, $\hat{\tau}$ is 1, if they are all discordant $\hat{\tau}$ is -1. *C* is the number of concordant pairs within ½n(n-1) pairs, 2*C*/*n*(*n*-1) can represent probability $\pi_c$ of ($X_i$, $Y_i$) and ($X_j$, $Y_j$) are concordant, 2*D*/*n*(*n*-1) can represents the probability $\pi_d$ of a pair is discordant.

Because $\pi_c$- $\pi_d = \hat{\tau}$, and $\pi_c + \pi_d = 1$, we can get another useful equation
$\pi_c/\pi_d = (1+\tau)/(1-\tau)$. (**Equation 6**)

In Example1:
Movie-Id: (Person1’s vote, Person2’s vote);
Movie 2: (1, 0.6); Movie 19: (0.6, 0.2);
Movie: 24 (0.8, 0.6); Movie: 32 (0.8, 0.8);
Movie 62: (1, 0.8); Movie 65: (0.8, 0)

We view 0 and 0 a concordant pair, and view 0 and the other non-zero numbers as neither discordant nor concordant, we just ignored them.

| movieId | person 1 | person 2 | |
|---|---|---|---|
| (2, 19) | 1-0.6=0.4 | 0.6-0.2=0.4 | Con |
| (2, 24) | 1-0.8=0.2 | 0.6-0.6=0 | |
| (2, 32) | 1-0.8=0.2 | 0.6-0.8=-0.2 | Dis |
| (2, 62) | 1-1=0 | 0.6-0.8=-0.2 | |
| (2, 65) | 1-0.8=0.2 | 0.6-0=0.6 | Cont |
| (19, 24) | 0.6-0.8=-0.2 | 0.2-0.6=-0.4 | Con |
| (19, 32) | 0.6-0.8=-0.2 | 0.2-0.8=-0.6 | Con |
| (19, 62) | 0.6-1=-0.4 | 0.2-0.8=-0.6 | Con |
| (19, 65) | 0.6-0.8=-0.2 | 0.2-0=0.2 | Dis |
| (24, 32) | 0.8-0.8=0 | 0.6-0.8=-0.2 | |
| (24, 62) | 0.8-1=-0.2 | 0.6-0.8=-0.2 | Con |
| (24, 65) | 0.8-0.8=0 | 0.6-0=0.6 | |
| (32, 62) | 0.8-1=-0.2 | 0.8-0.8=0 | |
| (32, 65) | 0.8-0.8=0 | 0.8-0=0.8 | |
| (62, 65) | 1-0.8=0.2 | 0.8-0=0.8 | Con |

**Table 4**

From table 4 we get *C* = 7, *D* = 2; *S* =7-2=5;
$\hat{\tau}$ = (2*5)/6*(6-1)=0.3333333
$\pi_C/\pi_D = (1+\tau)/(1-\tau)$=2

That represents (the probability of concordant/the probability of discordant)=2, which means if they have seen n movies in common, 2n/3 movies will be concordant, n/3 of them will be discordant.

**Generate Recommendation Algorithm**

After the AIS has chosen 100 people (antibodies) who have similar preferences with the user (antigen) require recommendations, the CF will use the ( Equation 6 to calculate the predictions.

$$prediction = \frac{\sum_{i=1}^{100}\left(weight_i \times VoteScore_i\right)}{\sum_{i=1}^{100}\left(weight_i \times 1\right)}$$

( Equation 6 )
*$weight_i$ reprents the weight of the ith antibody,*
*$weight_i$ = $concentration_i$*, ( the ith antibody voted this movie),
*$weight_i$ =0,* ( the ith antibody did not vote this movie )
*$concentration_i$* represents the concentration of

*VoteScore$_i$* represents the score which the ith antibody voted this movie as.

We use the concentration as the weight to calculate the prediction, because it contains both the correlation of the antibody to the antigen and the correlation of the antibody to the other antibodies.

**The Data**

The data we use in this project is publicly available data, which is offered by the Compaq Research (formerly DEC Research) [7]. It contains 2811983 ratings entered by 72916 users for 1628 different movies, and it has been used in numerous CF publications.

**RESULTS**

We calculated the ignored percent of zero ones in 350 pairs of persons which are randomly chosen from the database when we used Kendall tau to calculate their affinities.

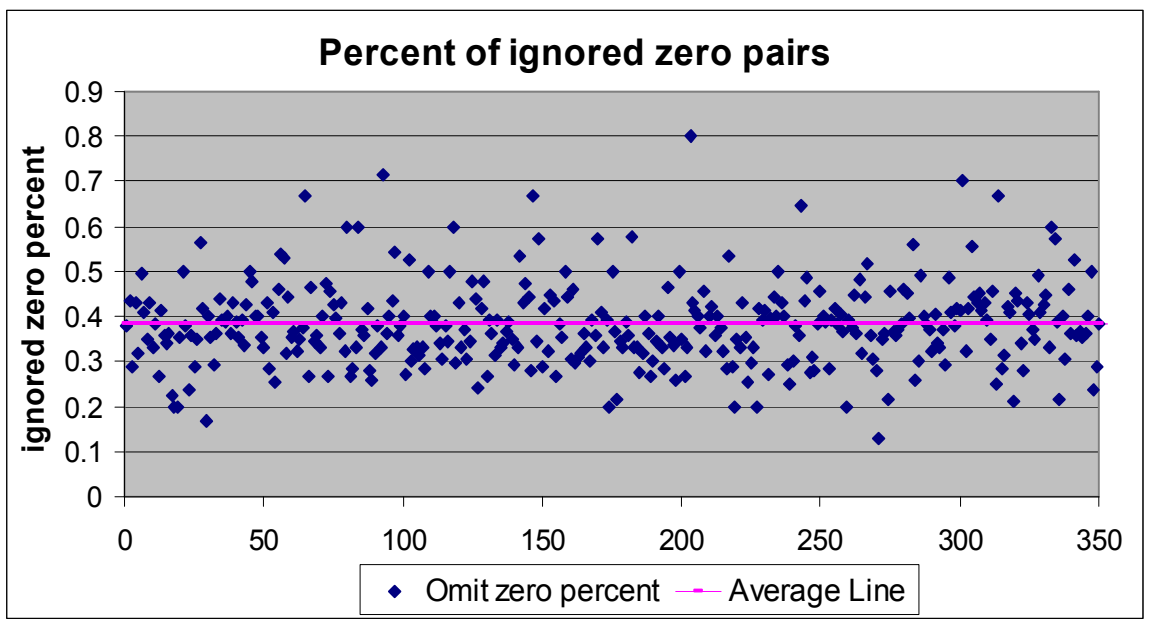

**Figure 2**( percent of Kendall tau ignored zero)

Figure 2 shows percent of ignored zero ones. In average, 38.27% information was ignored. In some occasions, more than 50% information was ignored.

In order to calculate the prediction accuracy we choose 300 persons (who voted more than 20 movies) randomly from the database, hide one of their votes for each person, offer the person prediction for the hidden movie using the information left. We do this 20 times for each person chosen by hiding different vote and compare these 20 predictions with their hidden actual votes.

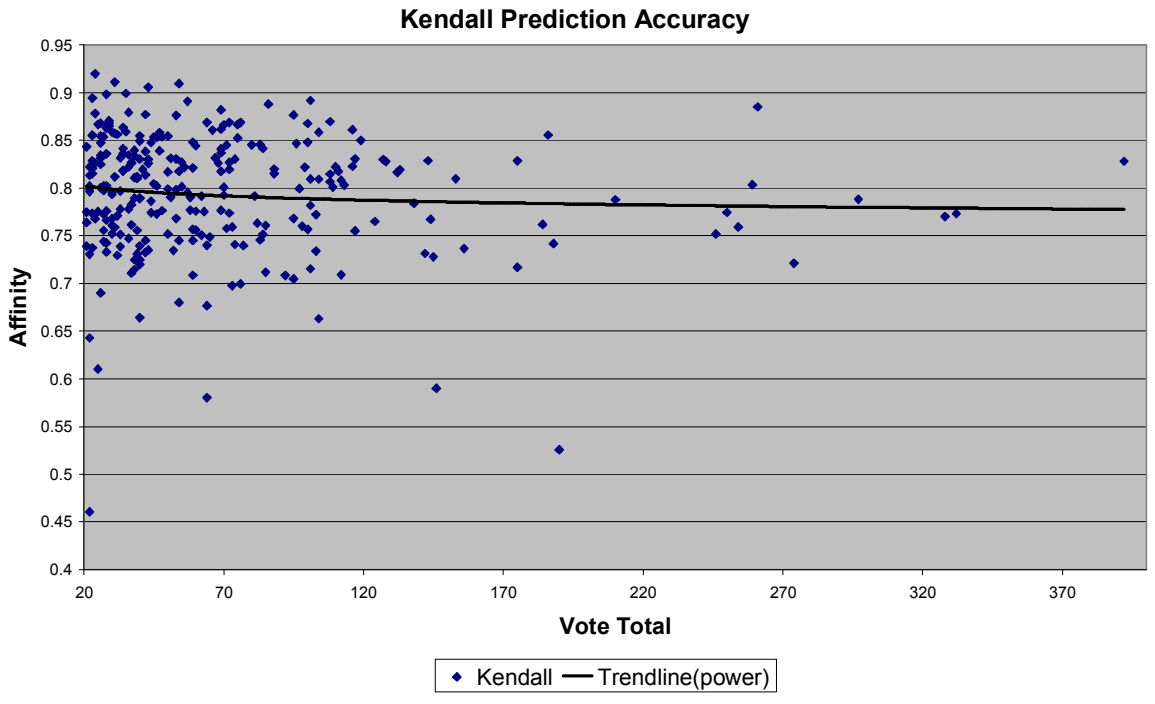

**Figure 3** (Prediction accuracy for Kendall)

Figure 3 shows the prediction accuracy for 300 users using the movie recommendation system which uses the Kendall tau correlation method. The mean prediction accuracy is 0.796419. The prediction accuracy is calculated using Equation 6.

$$PredictionAccuracy = 1 - \frac{|prediction - actualVote|}{20}$$

**Equation 7**

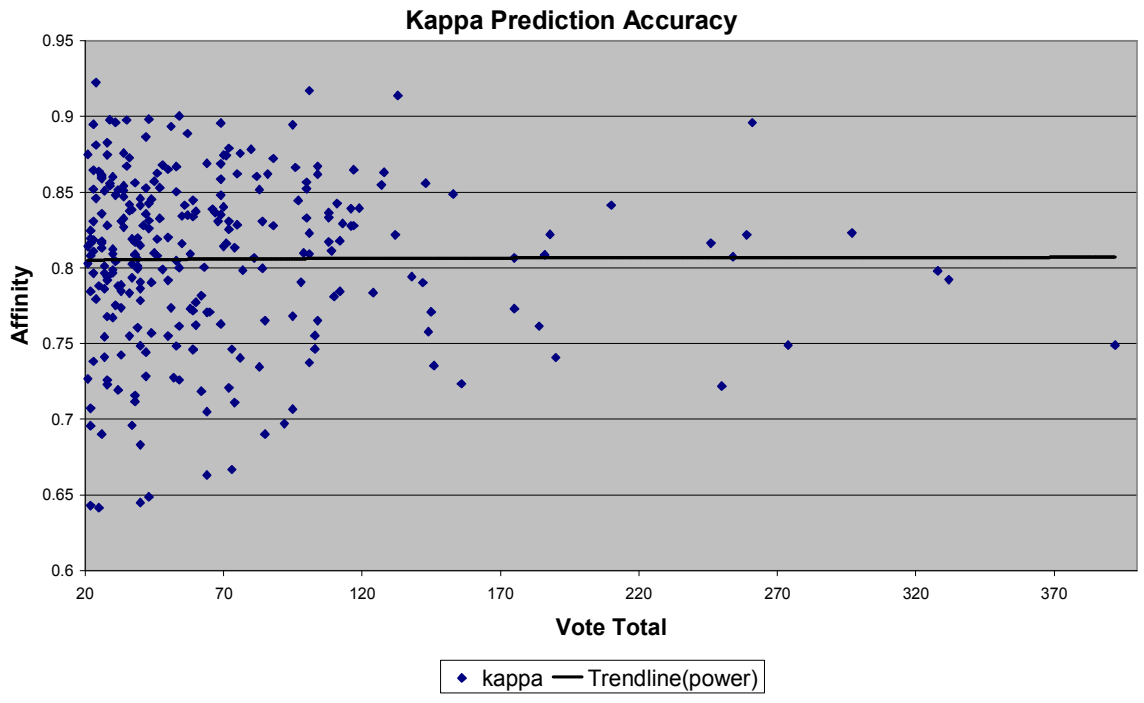

**Figure 4** (Prediction accuracy for Kappa)

Figure 4 Figure 4 shows the prediction accuracy for 300 users using the movie recommendation system which uses the Weighted Kappa correlation method The Mean prediction accuracy is 0.80762. The accuracy 0.8 represents that there is one rank difference between the prediction and the user's actual rating for a movie. So if the system predicts a movie as 'Very Good', the user may think that it is 'Good'.

We use the Weighted Kappa method to get the 100 antibodies for one user (the Antigen) and use the Kendall tau method to calculate the correlations between the 100 antibodies and the antigens. We compare the correlations got by Kappa and Kendall. The results are shown in Figure 5.

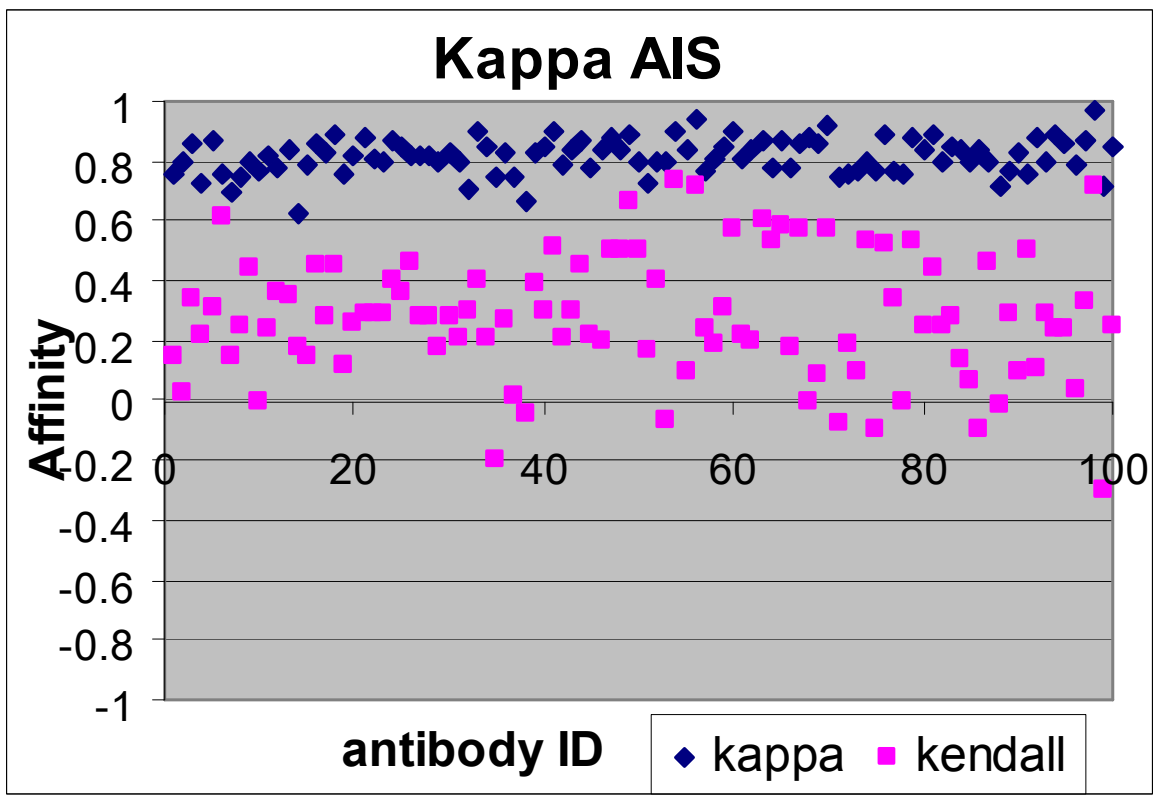

**Figure 5 (Kappa AIS)**

From Figure 5 we can see that all the Kappa values are very high (So we use kappa to implement the AIS) and some of the Kendall values are below zero.

We use the system which uses the Kendall tau method, to get the 100 antibodies for one user (the Antigen) and use the Weighted Kappa to method to calculate the correlations between the 100 antibodies and the antigens. We compare the correlations got by Kappa and Kendall. The results are shown in Figure 6.

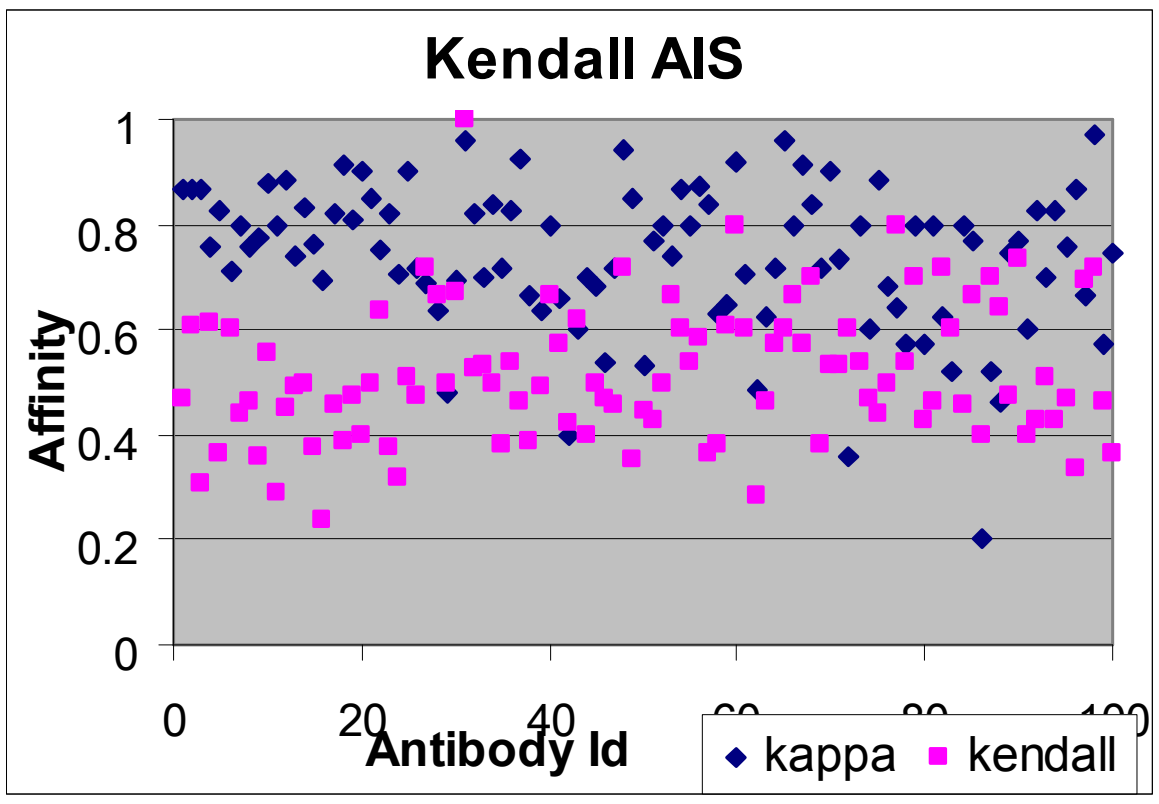

**Figure 6 (Kendall AIS)**

From Figure 6, we can see that sometimes Kendall is high (more than 0.2) but Kappa is not very high (less than o.6).

## DISCUSSION AND CONCLUSION

We found when we use the Kendall tau method in the movie recommendation systems, if we do not ignore these pairs, we treat the zero as opposite number or negative number, the Kendall tau value will be different we calculate it in different orders. Obviously it is wrong. If we ignore the pairs with zero, too much information we ignore. We think the problem is that there are only 6 categories, but more observations (for n observations there will be $\frac{1}{2} * n(n-1)$ comparisons) for the movie problems. When we use Kendall tau to calculate correlations, there will be many tires. In our opinions, the Kendall tau is not suitable for movies problems; Kappa is a good method for movies problems.

For this reason, if we use Kendall tau, the recommendation system should get worse results. But the prediction accuracy using these two methods has a slight difference. We have not known why this happen yet.